\pdfoutput=1

\documentclass[11pt]{article}

\usepackage{acl}

\usepackage{times}
\usepackage{latexsym}
\usepackage{multirow}
\usepackage{tikz}
\usetikzlibrary{positioning, shapes}
\usepackage{algorithm}
\usepackage{algorithmic}
\usepackage{xcolor}

\usepackage[T1]{fontenc}

\usepackage[utf8]{inputenc}

\usepackage{microtype}

\usepackage{inconsolata}
\usepackage{graphicx}
\usepackage{amsmath}
\usepackage{amssymb}
\usepackage{booktabs}
\usepackage{colortbl}
\definecolor{lightgray}{gray}{0.9}

 \usepackage{multirow}
\usepackage{amsfonts}
\usepackage{colortbl}
\usepackage{tcolorbox}
\usepackage{booktabs} 
\usepackage{tabularx}
\usepackage{framed}
\usepackage{arydshln}
\usepackage{subcaption}
\usepackage{arydshln}
\usepackage{xspace}
\usepackage{authblk}

\newcommand{\method}{\textsc{Struct-X}\xspace}
 
\title{\method: Enhancing Large Language Models Reasoning with Structured Data}

\author[$^{\heartsuit \dag}$]{Xiaoyu Tan}
\author[$^{\diamondsuit \dag}$]{Haoyu Wang}
\author[$^{*\diamondsuit \dag}$]{Xihe Qiu}
\author[$^\spadesuit$]{Yuan Cheng}
\author[$^\spadesuit$]{Yinghui Xu}
\author[$^{\heartsuit}$]{Wei Chu}
\author[$^\spadesuit$]{Yuan Qi}
\affil[$^{\diamondsuit}$]{Shanghai University of Engineering Science, Shanghai, China}
\affil[$^{\heartsuit}$]{INF Technology(Shanghai) Co., Ltd., Shanghai, China}
\affil[$^\spadesuit$]{Fudan University, Shanghai, China}
\affil[*]{This is to indicate the corresponding author}
\affil[$\dag$]{This is to indicate the equal contribution}

\begin{document}
\maketitle

\begin{abstract}
Structured data, rich in logical and relational information, has the potential to enhance the reasoning abilities of large language models (LLMs). Still, its integration poses a challenge due to the risk of overwhelming LLMs with excessive tokens and irrelevant context information. To address this, we propose \method, a novel framework that operates through five key phases: ``\textit{read-model-fill-reflect-reason}'' efficiently enabling LLMs to utilize structured data. It begins by encoding structured data into a topological space using graph embeddings, followed by filling in missing entity information with knowledge retrieval modules, and filtering out irrelevant tokens via a self-supervised module. The final phase involves constructing a topological network with selected tokens to further reduce the total token length for more effective LLM inference. Additionally, \method includes an Auxiliary Module trained to generate prompts, aiding LLMs in analyzing structured data. Extensive experiments on benchmarks, including the knowledge graph question-answer task and the long document reading comprehension task, show that \method notably improves LLM reasoning, demonstrating the effectiveness of structured data augmentation in improving LLM inference with complex input context. The code has been open-sourced and can be found in Appendix \ref{code}.
\end{abstract}

\section{Introduction}

In recent years, significant advancements have been made in the field of large language models (LLMs), particularly in natural language understanding \cite{fan2023large}. This progress has been largely driven by extensive pre-training on vast text corpora \cite{gao2023rarr}, which has enhanced their generation capabilities. These advancements are often viewed as critical steps towards the development of artificial general intelligence (AGI) \cite{pei2019towards}. During the deployment of LLMs as general-purpose assistants for a variety of real-world applications, it becomes necessary for LLMs to process multimodal inputs. Among these inputs, structured data, like structured knowledge graphs (KGs), is particularly important \cite{Ryen2022}. These graphs, with their rich repository of entity relationships and hierarchical knowledge, have the potential to significantly enhance the reasoning capabilities of LLMs, leading to more precise and reliable inferences.
However, in real-world applications, the effective utilization of structured knowledge in LLMs presents a significant challenge \cite{pan2024unifying}. A common approach is to flatten the structured information into a lengthy text sequence before inputting it into LLMs \cite{Li2023}. However, this method often introduces an excessive amount of task-irrelevant context. Excess information can overwhelm the models, thereby impairing inference efficiency and accuracy \cite{han2024intelligent}. Additionally, it hinders the ability of LLMs to accurately comprehend and represent the complex knowledge embedded within structured data \cite{Zhou2023}.

To address this issue, various approaches have been explored. Some studies have focused on converting knowledge graph triples into textual statements \cite{Zhang2023}, while others have emphasized incorporating knowledge graph embeddings \cite{Chen2023}. Additionally, efforts are underway to embed knowledge graph entities and relations directly into the encoder layers of LLMs \cite{Jiang2023}. More previous work is summarized in Appendix \ref{related}. However, these methods primarily concentrate on converting the structural data of knowledge graphs into different formats. They tend to overlook the need to reduce the information density of this structural data, which often includes task-irrelevant information. Moreover, these approaches face challenges in preserving the global topological structure of knowledge graphs, a critical aspect that warrants further attention.

In addition to the issues of redundant information and the lack of a global topological structure in knowledge graphs, another significant challenge is the high sparsity of these graphs  \cite{lazaridou2022internet}, characterized by missing semantic connections between entities. This sparsity presents a challenge for leveraging structural data in LLMs \cite{hadi2023survey}. LLMs tend to prioritize explicit semantic connections presented in the context while overlooking implicit connections, which are crucial for enhancing inference performance. Although current research, such as \cite{Lv2022} and \cite{Chai2023}, has been directed towards automatic knowledge completion and data augmentation to boost overall performance, these approaches tend to overlook the aforementioned challenges of redundancy and topological structure representation in utilizing structural data.

To overcome the existing bottlenecks discussed above, we introduce \method, a novel framework designed to utilize \textbf{\textit{Struct}}ured data to enhance the interaction and comple\textbf{\textit{X}} reasoning capabilities of LLMs. This framework is centered around a workflow of ``\textit{read-model-fill-reflect-reason}''. It employs the transformation of structured data into a topological space, achieved through the application of graph embeddings. This is followed by the augmentation of incomplete entity information utilizing knowledge retrieval modules. Subsequently, a self-retrieved generation module called Self-Reg is employed to eliminate irrelevant tokens. The final stage encompasses the development of a topological network incorporating the chosen tokens, which serves to diminish the overall token length, thereby enhancing the efficacy of LLM inference. Furthermore, an Auxiliary Module is also designed in \method, which adjusts prompts based on the loss, guiding the LLM generation. Extensive evaluation of knowledge graph QA and reading comprehension benchmarks have proven \method's superior reasoning abilities. These tests confirm that augmenting LLMs with structured data can significantly improve their inference skills in complex context environments. We refer interested readers to Appendix \ref{example} for more information about \method's interaction examples. The code of \method has also been open-sourced and can be found in Appendix \ref{code}. The main contributions of this paper include: 
\begin{enumerate}
    \item We propose a novelty framework \method that implements a process of ``\textit{read-model-fill-reflect-reason}'' on structured data, enabling LLMs to perform effective complex reasoning over structured data.
    \vspace{-1ex}
    \item We design a knowledge learning and filtering process to dynamically fill in structured knowledge gaps, coupled with a self-retrieved generation module called Self-Reg to filter and verify the relevance of retrieved knowledge, retaining valuable token information to alleviate learning burdens on LLMs.
        \vspace{-1ex}
    \item We construct specialized graph network encoders to fully learn the potential features of associated tokens and enable efficient cross-layer message passing in Transformers. We also devise an original Auxiliary Module for generating coherent prompts and improving answer responses. 
\end{enumerate}

\section{Preliminaries}

The task of text generation in LLMs involves creating a sequence of output \(y = [y_1, ..., y_T]\), where \(T\) represents the total number of tokens \cite{tang2023synthetic}, based on a given input prompt \(x\). This process is often modeled in an autoregressively manner, which estimates the likelihood of each token, where \(y_{<t}\) represents the tokens that come before the current sequence \([y_1, ..., y_{t-1}]\) \cite{Zhang2022}. Enhancements to this process can be made by incorporating relevant information from external documents \(D\) into the input, thereby refining the model's predictions \cite{Hofstatter2023}.
\begin{figure*}[ht]
     \centering
\includegraphics[width=1.02\textwidth, height=0.3\textwidth]{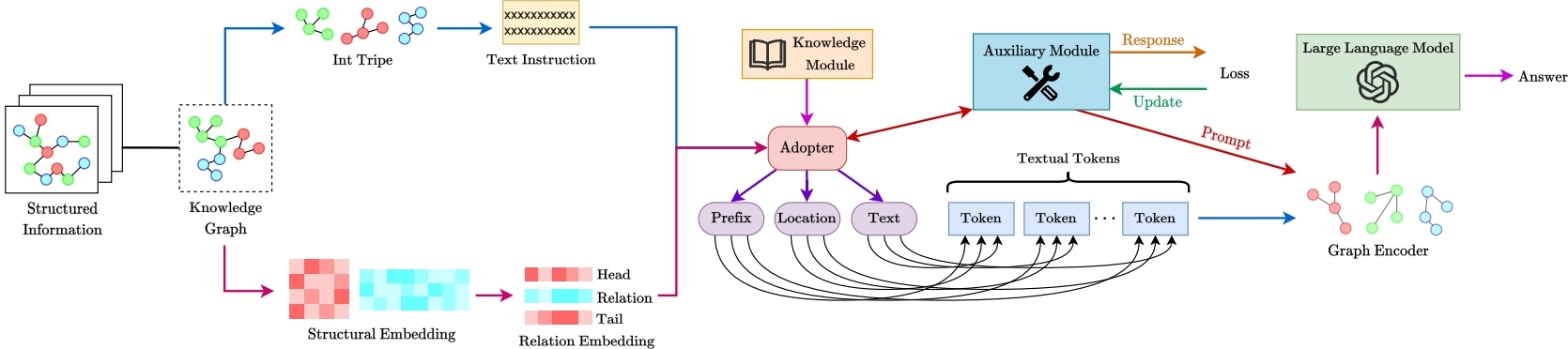} 
    \vspace{-4ex}
    \caption{Overall architecture of the proposed \method framework. It consists of modules for topological knowledge encoding, knowledge injection and retrieval, graph topology encoder, and Auxiliary Module.}
    \label{total}
    \vspace{-3ex}
\end{figure*}

Moreover, we can develop a novel decoding strategy that produces critique tokens \(\mathcal{C}\) alongside the main text output. These tokens are generated at each step and are designed to enable the LLMs to self-evaluate aspects such as relevance, factuality, and completeness of the generated content in Table \ref{t1}.
\vspace{-3ex}
\begin{equation}
p(y, \mathcal{C}|x) = \prod_{t=1}^{T} p(y_t, \mathcal{C}_t| x, y_{<t}, \mathcal{C}_{<t}),
\end{equation} where the critique token $\mathcal{C}_t$ depends on all preceding text and critiques. We define four types of critique tokens: \resizebox{!}{1.6ex}{\textcolor{blue}{\fbox{IFReT}}}  - predicts if retrieval is needed, \resizebox{!}{1.6ex}{\textcolor{orange}{\fbox{IFReL}}} - assesses passage relevance, \resizebox{!}{1.6ex}{\textcolor{brown}{\fbox{IFSuP}}} - checks output is supported and \resizebox{!}{1.6ex}{\textcolor{purple}{\fbox{IFUsE}}} - decides whether it is useful.

These critique tokens enable better control of the decoding process through re-ranking or constraints \cite{Asai2023}. For instance, the probability of a desirable \resizebox{!}{1.6ex}{\textcolor{orange}{\fbox{IFReL}}} token can upweight certain outputs. As an example, the attention distillation critique token is computed between the input $x$ and response $y_t$. This critiques the attention alignment between $x$ and $y_t$. By generating such reflective signals, \resizebox{!}{1.6ex}{\textcolor{blue}{\fbox{IFReT}}} can adapt its decoding strategy over time based on its critiques. The \resizebox{!}{1.6ex}{\textcolor{blue}{\fbox{IFReT}}} approach allows customization of model behavior through constraints on desired critique tokens. The detailed algorithm implementation process can be found in Appendix \ref{self}.

$p(\text{\resizebox{!}{1.6ex}{\textcolor{blue}{\fbox{IFReT}}}} = y | x) = f_{\phi}(x)$, which predicts whether passage retrieval is needed ($y$) given the input $x$ using a scoring function $f_{\phi}$ parameterized by $\phi$. The relevance scoring between a passage $p$ and the input is $s_{rel} = g_{\theta}(x, p) \cdot \text{\resizebox{!}{1.6ex}{\textcolor{orange}{\fbox{IFReL}}}}(x, p)$, where $g_{\theta}$ produces a relevance score modulated by the \resizebox{!}{1.6ex}{\textcolor{orange}{\fbox{IFReL}}} gate value. The factual consistency between a response $y$ and passage $p$ is evaluated by $s_{con} = h_{\psi}(y, p) \odot \sigma(\text{\resizebox{!}{1.6ex}{\textcolor{brown}{\fbox{IFSuP}}}}(y, p))$, where $\odot$ is element-wise production and $\sigma$ is the sigmoid activation function. The overall utility is decided using $ u = \text{\resizebox{!}{1.6ex}{\textcolor{purple}{\fbox{IFUsE}}}}(x, y)$.

\vspace{-1ex}
\section{Methods}

\vspace{-1ex}
\subsection{Topological Knowledge Injection}
We first implement ``\textit{read-model-fill}'' process and 
we start by processing input KGs using a graph attention encoder (GAE) that consists of $L$ layers \cite{xu2021graph}. The initial node features, denoted as $h_v^{(0)}$, are set up using information obtained from the KG completion module \cite{fei2021enriching}. After processing through $L$ layers, we obtain the final node embeddings, $h_v^{(L)}$, which effectively represent both the semantic and structural information of the KGs. These encoded graph embeddings, $h_v^{(L)}$, are then partially masked at a specific rate, $p_{mask}$, to assist in learning about missing knowledge. This masking process can be mathematically represented as $\Tilde{h}_v = M(h_v^{(L)})$, where $M(\cdot)$ symbolizes the masking operation. The masked nodes, denoted as $\Tilde{h}_v$, are then fed into the knowledge retrieval module, $R(\Tilde{h}_v)$, which is explained in the following section. This module plays a crucial role in supplementing the missing information, thereby facilitating the generation of complete graph embeddings $\Bar{h}_v$ \cite{Reda2022}. 
\begin{figure}[ht]
     \centering
\includegraphics[width=0.5\textwidth]{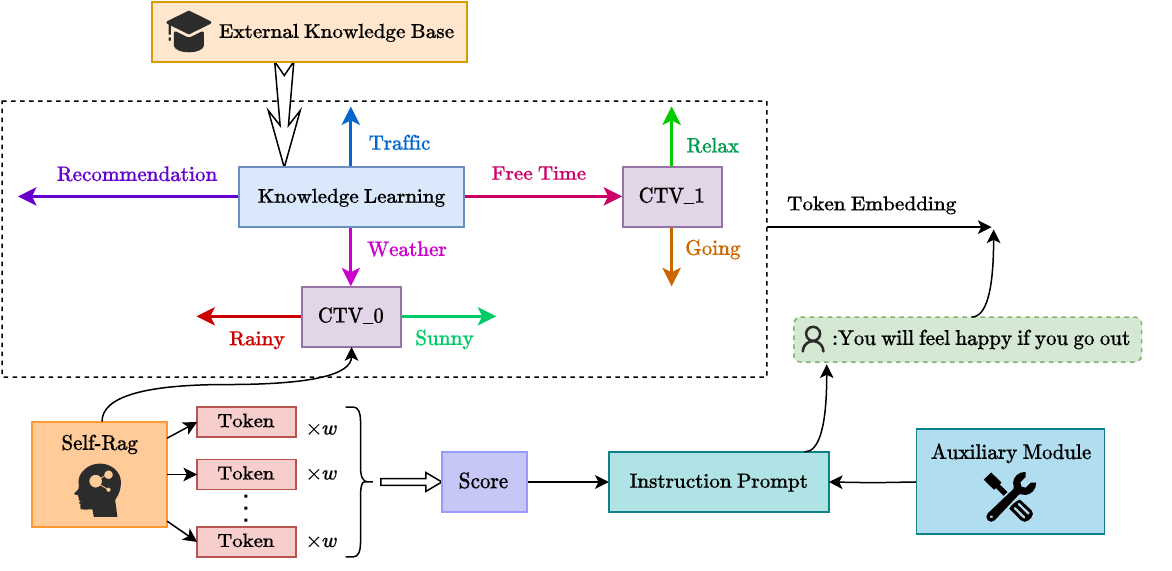}  
        \vspace{-4ex}
    \caption{Knowledge injection and retrieval modules in \method. The knowledge retrieval module fills in missing entity information in the graph embeddings.}
    \label{IR}
    \vspace{-2ex}
\end{figure}

To address the gaps in entity information within the structured knowledge graph, we have developed a knowledge learning module, denoted as $F$. This module is designed to retrieve pertinent facts from the knowledge base to enhance the masked node embeddings, $\tilde{h}_v$ \cite{yasunaga2022deep}. More specifically, for each masked node, we calculate a similarity score between its embedding $\tilde{h}_v$ and all tail entities $t$ that are part of the set $E$. This is achieved using the scoring function $f_{score}$, which can be represented as:
\begin{equation}
s(v,t) = f_{score}(\tilde{h}_v, t).
\end{equation} 
This process enables us to efficiently fill in the missing information in the knowledge graph.
\begin{table*}[h]
\centering
\resizebox{\textwidth}{!}{
\begin{tabular}{|l|l|l|l|}
\hline
\arrayrulecolor{black}  
Type & Inputs & Outputs & Descriptions \\
\hline
\setlength{\fboxrule}{1.2pt}  
\resizebox{!}{1.8ex}{\textcolor{blue}{\fbox{IFReT}}} & $query, context$ & $\{activate, wait\}$ & Decides when extra facts can assist reasoning \\
\hdashline
\setlength{\fboxrule}{1.2pt} 
\resizebox{!}{1.8ex}{\textcolor{orange}{\fbox{IFReL}}} & $query, evidence$ & $\{high, low\}$ & Whether evidence provides useful clues to solve query \\
\hdashline
\setlength{\fboxrule}{1.2pt}
\resizebox{!}{1.8ex}{\textcolor{brown}{\fbox{IFSuP}}} & $query, evidence, response$ & $\{strong, medium, weak\}$ & Alignment between statements in response and evidence \\
\hdashline
\arrayrulecolor{black}  
\setlength{\fboxrule}{1.2pt} 
\resizebox{!}{1.8ex}{\textcolor{purple}{\fbox{IFUsE}}} & $query, response$ & $\{5,4,3,2,1\}$ & Usefulness score of response in answering query \\
\hline
\end{tabular}
}
\caption{Self-supervised Auxiliary Module related parameters for selective knowledge retrieval and response correlation verification}
\label{t1}
\vspace{-3ex}
\end{table*}

The scoring function evaluates both feature and topological similarities within the graph in Figure \ref{IR}. It selects the top $K$ entities $t$ based on the highest scores and retrieves the related facts $(h, r, t)$ from the knowledge base. To incorporate these facts into the node embeddings, a relation-aware aggregation function $f_{agg}$ is used. This function accumulates relevant knowledge for each node, using a score threshold $\tau$ to filter out irrelevant facts \cite{yu2022jaket}. The aggregation function adeptly manages various relations in structured knowledge by considering information from retrieved triples in a relation-aware manner. Additionally, before being input into the Transformer encoder, one linear layer $o$ concatenates and processes embeddings from all GAT layers.
\vspace{-1ex}
\begin{equation}
\begin{aligned}
\bar{h}_v &= f_{agg}(\{\tilde{h}_v\} \cup \{(h, r, t) | s(v,t) > \tau\}),\\
e_v &= o([h_v^{(1)}, ..., h_v^{(L)}, \Bar{h}_v]) 
\end{aligned}
\label{relev}
\end{equation}

The $o$ merges inputs and reduces dimensionality. This process retains rich multi-scale structural and semantic features at various depths. The output $e_v$ is flattened and prepared into sequences to replace token embeddings for the Transformer encoder input, as suggested by \cite{Wang2021}. The refined node embeddings $\bar{h}_v$, enriched with retrieved entity information, supply additional knowledge for reasoning in the downstream LLMs. 

\subsection{Knowledge and Information Retrieval} 
\label{sec:reflect}
Here we perform the ``\textit{reflect}'' process. The module $R(\Tilde{h}_v)$ retrieves relevant knowledge absent in masked graph node inputs $\Tilde{h}_v$. Related entities can be dynamically discovered by matching tail entities $t$ to each masked node using a similarity scoring function $f_{score}(\Tilde{h}_v, t)$ considering both feature and topological similarity \cite{Lewis2020}. Related facts $(h, r, t)$ are recalled to fill gaps. After concatenating retrieved knowledge sequences for all nodes ordered by similarity scores, we employ a pruning algorithm leveraging multi-head self-attention distillation and thresholds to filter out lower-weighted tokens. The remaining dense sequence provides supplemental external knowledge to complete masked graph node inputs.

To filter and verify the relevance of retrieved knowledge, we design a self-retrieved generation module $\text{$SelfReg$}_{\psi}(k)$ parameterized by $\psi$ that takes as input the retrieved knowledge sequences $k$ and outputs a filtered subset $\hat{k}$ containing only the most valuable tokens \cite{Shuster2021}. Specifically, $\text{$SelfReg$}_{\psi}$ first encodes the knowledge sequence $k = (x_1, x_2, \ldots, x_N)$ using a Transformer encoder to obtain representations $h_i = f_{\text{$enc$}}(x_i)$. Next, we compute an importance score for each token $s_i = \sigma(f_{\text{score}}(h_i))$, where $f_{\text{$score$}}$ is a scoring network and $\sigma$ is a sigmoid activation function. To train the scoring network in a self-supervised manner, we create corrupted knowledge sequences $\tilde{k}$ by randomly masking or shuffling some tokens. A contrastive loss is implemented to assign higher scores $s_i$ to tokens from the original $k$ versus corrupted $\tilde{k}$:  
\vspace{-2ex}
\begin{equation}
\mathcal{L}_{\text{$contrast$}} = \sum_{i} \max(0, s_i - \tilde{s_i} + \Delta),
\end{equation}
where $\Delta$ is a margin hyperparameter. This drives the model to identify the most valuable knowledge. Finally, we filter the sequence by discarding tokens scoring below a threshold of $\tau$ to retain only the most relevant phrases, significantly reducing the learning burden when provided as supplements to the LLMs.
\vspace{-1ex}
\begin{equation}
\hat{k} = \{x_i | s_i > \tau\}.
\end{equation} 
The filtered relevant knowledge $\hat{k}$ provides targeted assistance to improve reasoning without overwhelming the LLMs with extraneous and irrelevant information.

\subsection{Graph Topology Encoder}
\begin{figure}[h]
     \centering
    \includegraphics[width=0.49\textwidth]{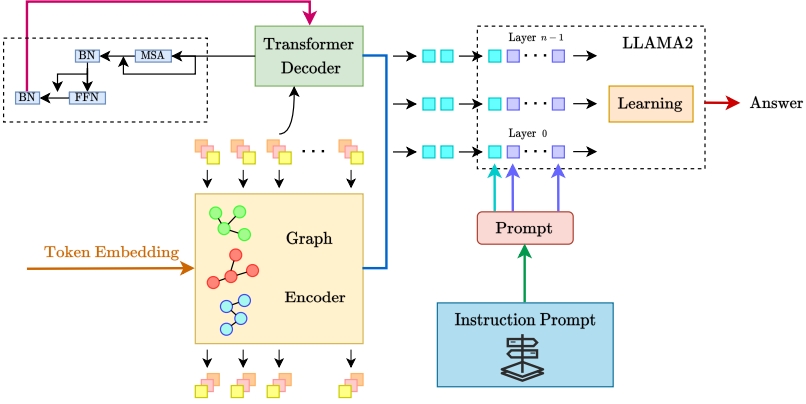}  
    \caption{Interaction between the graph topology encoder and LLM in \method. The encoder refines node embeddings via cross-layer message passing. The condensed embeddings are provided as supplements.}
    \label{TL}
    \vspace{-3ex}
\end{figure}

Here we perform the ``\textit{reason}'' phase. To capture semantic and structural interactions between entities within the KGs, we use a specialized graph encoder in Figure \ref{TL}, denoted as $E_{\theta}(G)$, which is parameterized by $\theta$. This KG is represented as $G=(V,E)$, where $V$ is the set of node entities and $E$ is the set of relation edges \cite{Li2022}. For each entity node $v_i$ in $V$, we first derive its initial feature representation $h_{v_i}^{(0)}$, which is a vector in a high-dimensional space.

The graph encoder works through a series of $L$ layers, each layer enhancing the node representations through message passing. This process can be described as:
\begin{equation}
h_{v_i}^{(l+1)} = f_{\theta}\left(\left\{h_{v_j}^{(l)}: v_j \in \mathcal{N}(v_i)\right\}\right), \quad \forall v_i \in V,
\end{equation} where $\mathcal{N}(v_i)$ refers to the neighboring nodes of $v_i$, and $f_{\theta}(\cdot)$ is a function that aggregates information from these neighbors to update the node's embedding.

To focus on the most relevant semantic connections, we use a graph self-attention layer within $f_{\theta}(\cdot)$. This layer calculates attention weights as follows $a_{ij} = \frac{\exp(\langle q_i, k_j \rangle)}{\sum_{v_k \in \mathcal{N}(v_i)} \exp(\langle q_i, k_k \rangle)}$, where $q_i$ and $k_j$ are derived from the embeddings of the nodes. This method allows the model to selectively emphasize the most informative signals from neighboring nodes \cite{cui2020adaptive}.

After processing through $L$ layers, we obtain refined node embeddings $z_{v_i} = h_{v_i}^{(L)}$, which encapsulate both semantic and structural information of the graph. To make these embeddings more manageable for downstream tasks, we compress them through a trainable down-projection layer:
\begin{equation}
e_{v_i} = \mathbf{W}_d z_{v_i}, \quad \mathbf{W}_d \in \mathbb{R}^{d_h \times d_e}, d_e < d_z.
\end{equation} This step reduces the dimensionality of the embeddings to $d_e$, which is smaller than the original $d_z$. The resulting condensed embeddings $e_{v_i}$ still retain crucial token-level interactions but are more concise, making them better suited for training models for specific tasks. This approach ensures that while the size of the input sequence is significantly reduced, the essential semantic and structural features of the knowledge graph are preserved for subsequent reasoning.

\vspace{-1ex}
\subsection{Auxiliary Module}
To further guide the LLMs in effectively reasoning over the structured input with a knowledge graph, we have developed an Auxiliary Module. This module is designed to create dynamic prompts that enhance the coherence of answers generated by LLMs. It functions by analyzing the LLM's predicted answer, denoted as \( \hat{y} \), along with the current loss, \( L \). Based on these inputs, it generates a refined prompt, \( p' \), which is then used for a new round of inference. We use the pre-trained Bert model (i.e., bert-base-NER) \cite{DBLP:journals/corr/abs-1810-04805} to construct this Auxiliary Module, symbolized as \( G \) and parameterized by \( \theta_g \). This generator crafts the prompt text, taking into account the input values $p' = G(L, \hat{y}; \theta_g)$. The generator is trained jointly with the overall system using policy gradient methods to maximize the expected reward $R$ of producing coherent answers:
$J(\theta_g) = \mathbb{E}{p' \sim G}[R(p')]$, $\nabla{\theta_g} J(\theta_g) = \mathbb{E}{p' \sim G}[\nabla{\theta_g}\log G(p'|L, \hat{y};\theta_g)R(p')]$.

This reward function is designed to encourage the LLMs to generate responses that are not only fluent but also logically consistent, particularly when using the updated prompt. This feature enables the module to dynamically adjust prompts based on the current performance, thereby offering new approaches to improve the quality of answers. Throughout the training process, the module progressively learns to produce more effective prompts, leading to enhanced accuracy and coherence in the LLM's reasoning. For more detailed experimental testing and analysis, please refer to Appendix \ref{Auxiliary}.

\section{Experiment}
\subsection{Datasets and Tasks}
\vspace{-1ex}
We assess the performance of our proposed \method framework on four open-source benchmark datasets designed for knowledge graph reasoning and multi-hop reasoning abilities on graphs. 

\textbf{Task1:WebQSP} contains 4,737 QA pairs where the questions require logical reasoning over a knowledge graph derived from Wikipedia to infer the correct answer. The knowledge graph consists of 5,719 entities and 2,150 relations. 

\textbf{Task2:MetaQA} comprises a set of more complex compositional questions constructed from an underlying knowledge graph with a vocabulary of 300 entities and 100 relations. It has a total of 1,200 unique questions that test the multi-hop, logical, and comparative reasoning abilities of models.

\textbf{Task3:Family Tree Age} 
Consider a family tree \( G = (V, E) \), where each individual \( v_i \) in \( V \) is associated with a description \( d_i \) specifying their age. The objective of this task is to identify the triplet comprising an individual, one of their grandparents, and a grand-uncle/grand-aunt by marriage that collectively has the highest cumulative age. 

\textbf{Task4:Travel Route Optimization}
Let \( G = (V, E) \) be a graph representing connected cities, where each city \( v_i \) in \( V \) has a description \( d_i \) with the travel toll or tax. The LLM must plan the route from a source to a destination city that minimizes the total toll paid. 

For all datasets, we incorporate the encoded graph representations into Llama2, which has been pre-trained on BookCorpus and English Wikipedia. Appendix \ref{case} is a case analysis of the experimental results on the datasets.
\begin{table*}[ht]
\centering
\vspace{-2ex}
\rowcolors{2}{gray!15}{white}
\resizebox{\textwidth}{!}{
\begin{tabular}{l|c|c|c|c|c|c|c|c|c|c|c|c|c|c|c|c}
\hline \hline
\rowcolor{gray!20}
\textbf{Model/Datasets} & \multicolumn{4}{c|}{\textbf{WebQSP}} & \multicolumn{4}{c}{\textbf{MetaQA}} & \multicolumn{4}{c|}{\textbf{Family Tree}} & \multicolumn{4}{c}{\textbf{Travel Route}} \\
\cline{2-17}
\rowcolor{gray!20}
& \textbf{Acc.} & \textbf{Prec.} & \textbf{Rec.} & \textbf{F1} & \textbf{Acc.} & \textbf{Prec.} & \textbf{Rec.} & \textbf{F1}& \textbf{Acc.} & \textbf{Prec.} & \textbf{Rec.} & \textbf{F1}& \textbf{Acc.} & \textbf{Prec.} & \textbf{Rec.} & \textbf{F1} \\
\hline
\rowcolor{gray!40}
\multicolumn{17}{c}{\centering Embedding-based Model} \\
\hline
TransE  & 67.32 & 68.41 & 65.47 & 66.91 & 74.63 & 74.44 & 75.01 & 74.72  & 73.45 & 71.83 & 66.92 & 63.84 & 62.27 & 60.45 & 65.28 & 62.56  \\

DistMult  & 69.21 & 69.91 & 69.08 & 69.49 & 69.34 & 72.32 &  61.93 & 62.83 & 60.93 & 61.47 & 60.08 & 60.62 & 61.74 & 60.36 & 64.19 & 62.05  \\

EmbedKGQA & 63.26 & 72.38 & 74.33 & 73.35 & 70.17 & 70.38 & 69.57 & 69.97 & 65.62 & 64.31 & 68.94 & 65.92 & 64.95 & 63.01 & 68.64 & 65.37 \\

ComplEx  & 67.36 & 68.21 & 66.27 & 65.64 & 72.86 & 71.64 & 71.18 & 69.53 & 59.83 & 58.45 & 62.74 & 60.27 & 57.45 & 55.92 & 60.93 & 57.84 \\

RotatE  & \underline{74.55} & 72.68 & \underline{76.93} & \underline{74.77} & 78.19 & 78.16 & \textbf{78.22} & \textbf{77.19} & 67.63 & 66.28 & 65.79 & 64.37 & 62.64 & 63.01 & 61.83 & 62.27 \\

\hline
\rowcolor{gray!40}
\multicolumn{17}{c}{\centering Open-source LLM} \\
\hline

Llama2$_{7B}$ & 27.13 & 29.17 & 25.23 & 27.36 & 26.45 & 25.73 & 28.82 & 26.94 & 32.81 & 34.73 & 30.16 & 32.26 & 25.92 & 27.84 & 23.68 & 25.47 \\

Alpaca$_{7B}$ & 33.56 & 31.83 & 36.29 & 33.72 & 22.74 & 25.93 & 19.82 & 21.45 & 28.36 & 26.92 & 31.74 & 28.91 & 39.45 & 41.27 & 37.63 & 39.18 \\

Llama2$_{13B}$ & 29.83 & 28.73 & 31.91 & 39.95 & 35.92 & 36.19 & 36.84 & 35.91 & 31.45 & 33.82 & 39.27 & 31.28 & 36.37 & 38.62 & 34.93 & 36.45 \\

Alpaca$_{13B}$ & 38.37 & 40.92 & 36.74 & 38.45 & 42.83 & 33.62 & 31.74 & 32.56 & 40.83 & 32.76 & 38.84 & 40.37 & 37.92 & 35.74 & 40.83 & 37.94 \\

ChatGPT & 41.27 & 49.84 & 44.92 & 41.72 & 38.74 & 37.56 & 40.83 & 38.94 & 35.92 & 33.84 & 38.62 & 35.83 & 34.83 & 36.73 & 42.15 & 44.71 \\

\hline
\rowcolor{gray!40}
\multicolumn{17}{c}{\centering LLM-based Fine-tuning} \\
\hline

KG-LLaMA  & 42.45 & 43.28 & 40.39 & 41.37 & 46.28 & 45.94 & 47.36 & 46.47 & 49.74 & 48.36 & 42.64 & 40.27 & 48.56 & 50.28 & 45.74 & 47.83 \ \\
KG-Alpaca & 48.92 & 47.74 & 51.83 & 49.45 & 43.56 & 42.41 & 46.28 & 43.94 & 45.92 & 46.84 & 44.56 & 45.62 & 42.91 & 50.74 & 47.28 & 43.45 \\
\hline

KG-BERT & 56.28 & 55.94 & 57.36 & 56.47 & 70.92 & 69.74 & 63.82 & 61.45 & 63.27 & 61.19 & 68.29 & 64.23 & 74.32 & \underline{75.61} & \underline{74.45} & 72.37 \\

PKGC & 64.56 & 67.44 & 62.64 & 64.10 & 77.79 & \underline{76.92} & 74.27 & 73.49 & \underline{79.32} & \underline{78.41} & \underline{75.36} & \underline{74.91} & \underline{76.35} & 75.42 & 74.37 & \underline{76.91} \\

Vanilla IT & 65.62 & 69.86 & 66.29 & 65.54 & \underline{78.28} & 74.31 & 72.98 & 73.62 & 71.57 & 70.58 & 69.63 & 70.05 & 65.94 & 67.32 & 63.69 & 64.96 \\

KoPA & 72.48 & \underline{72.82} & 71.64 & 71.52 & 75.82 & 74.69 & 73.95 & 71.58 & 73.56 & 72.18 & 72.08 & 71.53 & 73.13 & 75.05 & 71.12 & 71.24 \\

\hline
\rowcolor{gray!20}
Ours(\method) & \textbf{75.13} & \textbf{73.40} & \textbf{77.25} & \textbf{75.29} & \textbf{79.63} & \textbf{78.27} & \underline{77.53} & \underline{76.61} & \textbf{82.68} & \textbf{82.95} & \textbf{79.34} & \textbf{78.92} & \textbf{81.69} & \textbf{81.53} & \textbf{78.62} & \textbf{78.04} \\
\hline \hline
\end{tabular}
}
\caption{Performance comparison across different datasets and Tasks}
\label{results}
\end{table*}

\begin{table}[ht]
\centering
\vspace{-2ex}
\resizebox{\columnwidth}{!}{
\begin{tabular}{lcccc}
\hline \hline 
Methods & WQSP & \begin{tabular}{c} 
QA \\ 1hop
\end{tabular} & \begin{tabular}{c} 
QA \\ 2hop
\end{tabular} & \begin{tabular}{c} 
QA \\ 3hop
\end{tabular} \\
\hline 
KV-Mem & 48.3 & 84.3 & 74.5 & 46.2 \\
GraftNet & 53.1 & 84.1 & 72.3 & 62.3 \\
EmbedKGQA & 63.3 & 85.3 & 74.1 & 79.8 \\
NSM & 69.6 & 84.6 & $\underline{89.1}$ & 88.7 \\
UniKGQA & $\underline{71.6}$ & 86.8 & 82.8 & 92.3 \\
StructGPT & 64.8 & $\underline{87.4}$ & 87.1 & $\underline{92.9}$ \\
\hline  \hline 
Ours & $\mathbf{75.1}$ & $\mathbf{91.3}$ & $\mathbf{92.7}$ & $\mathbf{93.8}$ \\
$\quad$ $w/o$ SI (struct inference) & 54.2 & 85.6 & 85.3 & 86.5 \\
$\quad$ $w/o$ KGP (KG perform better) & 56.4 & 86.9 & 87.5 & 89.8 \\
$\quad$ $w/o$ model (graph study) & 61.6 & 87.4 & 89.2 & 85.7 \\
$\quad$ $w/o$ encoder (location) & 69.3 & 88.1 & 76.6 & 79.3 \\
$\quad$ $w/o$ self-reg (correction) & 70.4 & 85.3 & 87.4 & 86.8 \\
\hline \hline 
\end{tabular}
}
\caption{Performance evaluation and comparison across different functional modules on reasoning tasks}
\label{q1}
\vspace{-3ex}
\end{table}

\subsection{Implementation Details}
\textbf{Baseline}
\begin{itemize}
\vspace{-1ex}
\item \textbf{Embedding-based Model}: We compare against representative embedding models for knowledge graph reasoning including TransE \cite{Bordes2013}, DistMult \cite{Yang2015}, EmbedKGQA \cite{Saxena2020}, ComplEx \cite{Trouillon2016}, and RotatE \cite{Sun2019}. 
\vspace{-1ex}

\item \textbf{Open-source LLM}: We evaluate reasoning capabilities of widely-used pre-trained language models accessible through open APIs, including Llama2 [7B \& 13B] \cite{touvron2023llama2} and Alpaca [7B \& 13B] \cite{yao2023exploring} which are openly available LLMs up to 13 billion parameters.
\vspace{-1ex}

\item \textbf{LLM-based Fine-tuning}: To assess the performance of LM fine-tuning approaches, we include as baselines KG-LlaMA and KG-Alpaca \cite{Yao2023}, KG-BERT \cite{Yao2019}, PKGC \cite{Lv2022} and vanilla IT \cite{Zhang2023} which incorporate techniques to enhance LMs using annotated KG datasets or self-supervision. 
\end{itemize}

Our implementation is in PyTorch and we run experiments on NVIDIA A100 GPUs. More details of training can be found in Appendix \ref{set}.

\subsection{Main Results}
The results presented in Table \ref{results} indicate that \method consistently outperforms existing baseline methods across various datasets. Specifically, in the WebQSP benchmark, \method achieves an accuracy of 75.13\%, which is 2.65\% higher than the previously best-performing method, KoPA. Additionally, \method shows modest improvements in precision and recall, with increases of 9.51\% and 10.96\%, respectively, compared to Vanilla IT. In the more challenging MetaQA dataset, \method's performance is notably better, surpassing the state-of-the-art accuracy scores by 1.84\% and achieving a 1.68\% higher precision. Furthermore, \method demonstrates significant advancements in specialized tasks such as Family Tree and Travel Route, where it exceeds the top baseline results by 3.36\% and 5.34\% in accuracy, respectively.


Compared to embedding models such as TransE, DistMult, and EmbedKGQA, \method also shows promising improvements in reasoning abilities by integrating both semantic and topological structures of knowledge graphs. For instance, against RotatE's accuracy of 74.55\% on the WebQSP dataset, \method achieves higher performance with a 75.13\% accuracy, an increase of 0.58\%. The difference is slightly more pronounced on the MetaQA dataset, where \method exceeds RotatE's score of 78.19\% by 1.44\% in accuracy. In scenarios requiring complex reasoning inferences, \method demonstrates enhanced capabilities, outperforming peak embedding model accuracy by a notable margin of 16.74\% in Task4.


The results also show that \method can enhance the capabilities of the Llama2, which itself achieves a 27.13\% accuracy on the WebQSP benchmark. This enhancement is achieved through masking graph embeddings and using topology matching to retrieve relevant facts, thus addressing the gaps in factual knowledge that Llama2 requires. By overcoming these deficiencies in LLMs, \method significantly improves performance, increasing accuracy by 47.4\%. This indicates the effectiveness of structured augmentation, which is not present in the Llama2. Further, \method filters out less important tokens using the Self-Reg module, ensuring focus on the most relevant information. In comparison to previous methods like KG-BERT fine-tuning, StructX offers essential enhancements, particularly in complex reasoning tasks, as evidenced by increases of up to 10.24\% in accuracy and 5.61\% in recall.

Based on the experimental results, the ``\textit{reflect}'' process also plays a crucial role in enhancing reasoning capabilities. This process involves the \text{\resizebox{!}{1.6ex}{\textcolor{blue}{\fbox{IFReT}}}}$(x)$, which selectively gathers evidence as needed, and the \text{\resizebox{!}{1.6ex}{\textcolor{orange}{\fbox{IFReL}}}}$(x,p)$, which filters less relevant passages using relevance scores from Eq.\ref{relev} to improve context for LLMs. Additionally, the \text{\resizebox{!}{1.6ex}{\textcolor{brown}{\fbox{IFSuP}}}}$(y,p)$ and \text{\resizebox{!}{1.6ex}{\textcolor{purple}{\fbox{IFUsE}}}}$(x,y)$ ensure passage-response consistency and assess overall utility, contributing to higher quality results. For further case studies and experiments on this topic, readers are directed to Appendix \ref{see}.

\vspace{-1ex}
\subsection{Ablation Study}
\subsubsection{Q1: Different functional modules}
Table \ref{q1} shows that each component of \method plays a crucial role in enhancing various reasoning capabilities. For 1-hop single fact questions, while all versions of \method are effective, the complete model excels with a 91.3\% accuracy due to its ability to perform combinatorial reasoning using multi-head attention. This is key for interpreting semantic connections. In 2-hop and 3-hop multi-step reasoning, the absence of knowledge retrieval and injection modules results in a significant performance drop, with decreases of 7.4\% and 7.3\% respectively. However, the full \method model, utilizing these modules, reaches 92.7\% and 93.8\% accuracy by effectively traversing distant nodes. The graph topology encoder also proves vital; its omission leads to a 5.8\% decline in location-based reasoning, highlighting its importance in connecting nodes and facilitating spatial/hierarchical reasoning through message passing. Furthermore, the lower accuracy without the Auxiliary Module underlines its utility in guiding coherent inference across multiple steps.
\vspace{-1ex}
\begin{table*}[h]
\centering
\rowcolors{2}{}{gray!25}
\resizebox{0.8\linewidth}{!}{
\vspace{-1ex}
\begin{tabular}{l@{\hskip 0.3cm}c@{\hskip 0.5cm}c@{\hskip 0.5cm}c@{\hskip 0.5cm}c@{\hskip 0.5cm}c@{\hskip 0.5cm}c}
\toprule
\multirow{2}{*}{\textbf{Model Variant}} & \multicolumn{3}{c}{\textbf{WebQSP}} & \multicolumn{3}{c}{\textbf{MetaQA}} \\
\cmidrule(lr){2-4} \cmidrule(l){5-7}
& Acc. (\%) & Rec. (\%) & F1 (\%) & Acc. (\%) & Rec. (\%) & F1 (\%) \\
\midrule
No Filtering & $71.2 \pm 1.41$ & $68.4 \pm 1.59$ & $69.8 \pm 1.47$ & $75.6 \pm 1.19$ & $72.3 \pm 1.38$ & $73.9 \pm 1.26$ \\
\rowcolor{gray!25}
Random (20\% cut) & $72.8 \pm 1.28$ & $70.1 \pm 1.44$ & $71.4 \pm 1.33$ & $77.2 \pm 1.07$ & $74.2 \pm 1.24$ & $75.6 \pm 1.13$ \\
Random (40\% cut) & $73.6 \pm 1.21$ & $71.3 \pm 1.36$ & $72.6 \pm 1.26$ & $78.1 \pm 1.01$ & $75.4 \pm 1.17$ & $74.7 \pm 1.08$ \\
\rowcolor{gray!25}
Random (60\% cut) & $74.1 \pm 1.16$ & $72.8 \pm 1.30$ & $73.3 \pm 1.21$ & $79.0 \pm 0.95$ & $76.8 \pm 1.11$ & $75.9 \pm 1.03$ \\
Self-Reg (20\% cut) & $74.9 \pm 1.09$ & $73.2 \pm 1.23$ & $74.0 \pm 1.14$ & $79.3 \pm 0.92$ & $78.2 \pm 1.06$ & $76.1 \pm 0.99$ \\
\rowcolor{gray!25}
Self-Reg (40\% cut) & $\textbf{75.1} \pm \textbf{1.07}$ & $\textbf{73.4} \pm \textbf{1.21}$ & $\textbf{75.3} \pm \textbf{1.12}$ & $\textbf{79.6} \pm \textbf{0.90}$ & $\textbf{78.3} \pm \textbf{1.04}$ & $\textbf{76.6} \pm \textbf{0.96}$ \\
Self-Reg (60\% cut) & $73.9 \pm 1.19$ & $72.7 \pm 1.34$ & $73.2 \pm 1.24$ & $78.9 \pm 0.97$ & $77.8 \pm 1.12$ & $76.4 \pm 1.03$ \\
\bottomrule
\end{tabular}}
\caption{Analysis of using different token filtering ratios in Self-Reg module}
\label{q4}
\vspace{-3ex}
\end{table*}

\subsubsection{Q2:Filtering and reflection mechanism}
Table \ref{q4} compares reasoning performance with the following variants: $StructX_{No Filtering}$: Directly injects all retrieved knowledge without filtering. $StructX_{Random Filtering}$: Randomly removes of retrieved tokens. $StructX_{Reg Filtering}$: Uses the proposed Self-Reg module to score and filter tokens.

Across WebQSP and MetaQA datasets, incorporating filtering mechanisms leads to consistent gains over no filtering baselines. Randomly removing tokens brings minor improvements, showing that some knowledge reduction is beneficial. However, learned filtering with Self-Reg leads to more substantial gains. Comparing different Self-Reg cutting ratios, 40\% filtering seems to achieve the optimal trade-off, maximizing accuracy and recall. More aggressive 60\% cutting starts to degrade performance likely due to removing pertinent facts. On the other hand, light 20\% filtering retains more distracting information. By balancing knowledge breadth and depth, 40\% Self-Reg filtering enhances language model inference without overwhelming models. By scoring and removing extraneous tokens based on contextual representations, Self-Reg retains the essence to augment language models without diverting attention.

\vspace{-1ex}
\subsubsection{Q3: Learning by Auxiliary Module}
The results in Table \ref{q2} demonstrate that incorporating the Auxiliary Module leads to significant performance gains over the base \method model without this component. We observe absolute improvements of 3.9\% in accuracy, 2.58\% in precision, and 5.72\% in recall after implementing the Auxiliary Module. This validates its efficacy in providing adaptive prompts that elicit more accurate and logically coherent reasoning from the LLM when inference is made over structured knowledge graphs. The gains over the previous best model, PKGC are also substantial, at 10.57\% higher accuracy. Hence, the auxiliary module proves important for multi-hop reasoning and steering deductions in the right direction over complex topological structures. The consistent benefits confirm that modeling explicit prompt-answering mechanisms customized for structured reasoning tasks is an effective approach.
\begin{table}[h]
\centering
\vspace{-2ex}
\resizebox{\columnwidth}{!}{
\begin{tabular}{l|c|c|c}
\hline \hline
Model & Accuracy (\%) & Precision (\%) & Recall (\%) \\ \hline
KG-BERT & $56.28 \pm 2.12$ & $55.94 \pm 2.35$ & $57.36 \pm 1.97$ \\
PKGC & $64.56 \pm 1.84$ & $67.44 \pm 1.62$ & $62.64 \pm 2.14$ \\
\method w/o Auxiliary Module & $71.23 \pm 1.27$ & $70.82 \pm 1.38$ & $71.53 \pm 1.22$ \\
\method & $\textbf{75.13} \pm \textbf{0.98}$ & $\textbf{73.40} \pm \textbf{1.12}$ & $\textbf{77.25} \pm \textbf{0.86}$ \\
\hline \hline
\end{tabular}
}
\caption{Performance from incorporating the Auxiliary Module for steering prompt}
\label{q2}
\vspace{-1ex}
\end{table}

\subsubsection{Q4:Knowledge injection variants}
To validate the contributions of different components of our knowledge injection mechanism, we conduct an ablation study with the following variants: $StructX_{No Injection}$: The base LLM (i.e., Llama2) without any graph representation injection. $StructX_{Embeddings Only}$: Encoded graph embeddings are directly injected without any masking or knowledge retrieval. $StructX_{Masking Only}$: Graph embeddings are masked but missing facts are not filled via retrieval. $StructX_{Retrieval Only}$: Masked embeddings are completed with the knowledge retrieval module but without graph encoding.
We compare reasoning performance on WebQSP and MetaQA benchmarks against these reduced injection variants. The results in Table \ref{q3} demonstrate clear improvements from collectively incorporating all knowledge injection components compared to ablated variants. The full \method model with topological encoding, masking, and retrieval achieves 1.68\% and 1.31\% higher accuracy over the best partial variant on WebQSP and MetaQA respectively. This confirms that each mechanism provides unique benefits - topological encoding better retain intricate connections, masking identifies missing facts, and retrieval fills knowledge gaps. The experiment proves that dynamic masking and retrieval to address inherent incompleteness in structured data are most impactful. Variants without these processes show worse performance as they fail to overcome language models' factual deficiencies.
\vspace{-1ex}
\begin{table}[ht]  
\centering
\vspace{-1ex}
\resizebox{\columnwidth}{!}{
\begin{tabular}{l|c|c} 
\hline \hline
Model Variant & WebQSP Accuracy (\%) & MetaQA Accuracy (\%) \\ \hline
No Injection & $63.45 \pm 1.72$ & $71.23 \pm 1.43$ \\
Embeddings Only & $68.92 \pm 1.37$ & $74.56 \pm 1.21$ \\
Masking Only & $71.23 \pm 1.19$ & $76.92 \pm 1.08$ \\
Retrieval Only & $73.45 \pm 1.04$ & $78.32 \pm 0.92$ \\
\textbf{\method} & $\textbf{75.13} \pm \textbf{0.98}$ & $\textbf{79.63} \pm \textbf{0.84}$ \\
\hline \hline
\end{tabular}
}
\caption{Advantages of dynamic factual injection and self-verified retrieval over individual knowledge supplementation variants}
\label{q3}
\vspace{-2ex}
\end{table}

\section{Conclusion}
\vspace{-1ex}
In this paper, we introduce \method, a groundbreaking framework designed to enhance LLMs in complex reasoning tasks. \method applies an efficient ``\textit{read-model-fill-reflect-reason}'' methodology to structured data. It is adept at learning graph embeddings that are sensitive to geometric contexts, capturing the content of entities as well as their topological relationships. This enables \method to effectively infer missing facts about entities by matching similar topological features. Furthermore, it enhances the LLMs by distributing multi-scale features, which bolsters the representation of underlying connections that are not explicitly apparent. \method excels in tasks such as knowledge graph-based QA tasks and reading comprehension, especially in scenarios that require multi-hop logical reasoning.

\vspace{-1ex}
\section{Limitiation}
\vspace{-1ex}
The knowledge graph encoding may not fully capture complex relationships beyond structural topology, the auxiliary module's prompting could be overly biased by the current loss landscape. Exploring more expressive graph representations and smarter prompting strategies could potentially address these limitations.

\bibliography{anthology,custom}

\clearpage

\appendix
\section{StructX Interaction Examples} \label{example}

\vspace{-1ex}
\begin{tcolorbox}[
  colback=white,
  colframe=black,
  sharp corners,
  title=Instructions,
  fonttitle=\bfseries
]
Please indicate whether referring to external documents, improves the quality of the generated response. Please respond with either [Yes] or [No] and provide a brief explanation.

\medskip

\textbf{Instruction}: Identify the shortest path between two nodes in this knowledge graph.

\textbf{Need retrieval?} \resizebox{!}{1.5ex}{\textcolor{orange}{\fbox{[Yes]}}}

\textbf{Explanation}: \method can ingest the graph structure and topology to reason about paths. \textcolor{orange}{But retrieving additional facts on edge distances or weights can supplement its understanding for more accurate optimization.}

\textbf{Instruction}: Determine which family tree node has the oldest relative based on date descriptions.  

\textbf{Need retrieval?} \resizebox{!}{1.5ex}{\textcolor{red}{\fbox{[No]}}}

\textbf{Explanation}: \method encodes the hierarchical tree relations and date informations directly \textcolor{red}{without needing external evidence. Retrieval may introduce unnecessary details.}

\textbf{Instruction}: Analyze the impacts of this new tax policy based on economic concepts.

\textbf{Need retrieval?} \resizebox{!}{1.5ex}{\textcolor{blue}{\fbox{[Yes]}}}

\textbf{Explanation}: While \method has some linguistic capabilities, \textcolor{blue}{retrieving domain knowledge on economics and regulations will improve understanding of entities and contextual impacts} for better analysis.

\textbf{Instruction}: Summarize the key events in this 5-page history passage.  

\textbf{Need retrieval?} \resizebox{!}{1.5ex}{\textcolor{green}{\fbox{[No]}}}  

\textbf{Explanation}: \method is designed to ingest long document passages directly through encoders. \textcolor{green}{No need for external info.}

\textbf{Instruction}: Compare the costs of different flight options based on stop, mileage and fare data.   

\textbf{Need retrieval?} \resizebox{!}{1.5ex}{\textcolor{brown}{\fbox{[No]}}}

\textbf{Explanation}: \method can encode and reason over structured data tables natively. \textcolor{brown}{External retrieval of similar data is unneeded.}
\label{ap1}
\end{tcolorbox}

\section{Related Work}\label{related}
Prior efforts have explored various techniques to enhance language models with structured knowledge. Pan \emph{et al.} \cite{Pan2024} directly converted knowledge graph triples into textual statements as inputs. Liu \emph{et al.} \cite{Liu2020} embedded knowledge graphs into hidden layers of BERT models. Harnoune \emph{et al.}\cite{Harnoune2021} utilized memory modules to provide facts for multi-hop reasoning. Wu \emph{et al.} \cite{Wu2020} designed graph recurrent networks to propagate information over knowledge graphs. Roux \emph{et al.}\cite{Roux2022} employed reinforcement learning for path-finding in knowledge-augmented environments. Tu \emph{et al.} \cite{Tu2021} used graph convolutional networks to distill subgraph knowledge. Liu \emph{et al.} \cite{Liu2022} pruned knowledge graph inputs via self-supervised learning objectives. Honovich \emph{et al.}\cite{Honovich2022} focused on improving factual consistency in conditional text generation. Shah \emph{et al.} \cite{Shah2020} iteratively expanded anchor sets for more comprehensive grounding. Xue \emph{et al.} \cite{Xue2020} optimized path selection on heterogeneous knowledge sources. While providing valuable techniques, existing methods lack consideration for efficiently filtering and condensing structured knowledge supplements tailored for language model integration. Our work aims to address these limitations through specialized contextual encoding, retrieval, and distillation components in the proposed \method framework.

\section{Case Study}\label{case}
\begin{tcolorbox}[colback=lightgray!20, boxrule=1pt, sharp corners, fontupper=\footnotesize]
\textbf{Family Relations}
\begin{flushleft}
John is Nikhil's paternal grandfather. 
Nikhil's aunt Meera is John's daughter.
John was born in 1950.
Nikhil was born in 2000.
Meera was born in 1975.\\

Raj is Nikhil’s maternal grandfather.  
Raj has a brother named Shekhar.
Raj was born in 1948.
Shekhar was born in 1946. \\

\textbf{Question:} What is the age difference between the oldest and youngest family members described?

\textbf{Response:} 
Based on the family tree details provided:

- The oldest member is Shekhar, Nikhil's grand uncle, born in 1946.  

- The youngest member is Nikhil, born in 2000.

- Their age difference is 1946 - 2000 = 54 years.

So the age difference between the oldest and youngest family members is \textbf{54 years}.  

\textbf{Correct Answer:}
54 years old.

\end{flushleft}
\label{ap2}
\end{tcolorbox}

In this family tree example, \method first encodes the hierarchical relations between members like ``John is Nikhil's paternal grandfather" and ``Meera is John's daughter" as structured knowledge. The connections allow tracing ancestry and descendent paths. Age details like ``John was born in 1950" are also injected. With encoded topology and dates, \method can traverse the tree to identify the oldest member, Shekhar and the youngest, Nikhil, before subtracting their years to find the 54-year difference. This showcases interpreting familial links and performing numerical reasoning over injected facts. By attending over neighborhood nodes, \method focuses on the most relevant semantic connections. The topology learning distinguishes close and distant relatives to assess ages. Retrieved date details fill knowledge gaps for calculation. Overall, this case validates \method's capabilities in encoding complex structure topology and performing multi-step inference by combining connection reasoning and data-driven deduction.

The example proves \method can encode intricate hierarchical structures and use encoded topology to trace relationships and inject valuable factual knowledge. By learning contextual representations and connections in structured data, \method successfully interprets semantic links between entities and integrates supplementary date details for numerical reasoning over multiple inference steps. This supports complex reasoning across topological dimensions.

\begin{algorithm}
\caption{Topology Learning and Training}
\label{ap41}
\begin{algorithmic}[1]
\STATE \textbf{Input:} knowledge graph $G = (V, E)$, LLM $M_{\theta}$, Auxiliary Module $A_{\phi}$
\STATE Encode $G$ into latent embeddings $Z_V$
\STATE Mask node embeddings at rate $p_{\text{mask}}$ as $\tilde{Z}_V$
\STATE \textbf{Step 1: Topology Modeling \& Filling}
\FOR{$v_i \in \tilde{Z}_V$}
\STATE Retrieve related facts $\mathcal{F}_i$ via similarity scoring
\STATE Update $\tilde{z}_i$ with $\mathcal{F}_i$ using $f_{\text{$agg$}}$
\ENDFOR
\STATE \textbf{Step 2: Graph Topology Reasoning}
\FOR{$l=1,\ldots,L$}
\STATE Message passing layer to update $Z_V$
\STATE Attention distillation over $G$
\ENDFOR
\STATE \textbf{Step 3: LLM Integration \& Training}
\STATE Flatten $Z_V$ and pack into sequences
\STATE Create auxiliary prompts $p_{\phi}$ with $A_{\phi}$
\STATE Jointly train $M_{\theta}$ on sequences using $p_{\phi}$
\STATE Update $A_{\phi}$ using policy gradient
\end{algorithmic}
\end{algorithm}

\begin{algorithm}[ht]
\caption{Knowledge Filtering Module}
\label{ap42}
\begin{algorithmic}[1]
\STATE {\bfseries Input:} $x$ (Input text)
\STATE {\bfseries Output:} $\{y, n\}$ (Retrieve), $\{r, ir\}$ (Relevant), $\{c, ic\}$ (Coherent)

\STATE Retrieve Module:
\STATE Decide whether passage retrieval is needed based on $x$
\STATE \quad {\bfseries if} retrieval is needed {\bfseries then}
\STATE \quad \quad Set $y$ to Yes
\STATE \quad \quad Set $n$ to No
\STATE \quad {\bfseries else}
\STATE \quad \quad Set $y$ to No
\STATE \quad \quad Set $n$ to No retrieval needed
\STATE \quad {\bfseries end if}

\STATE Relevant Module:
\STATE \quad Filter out irrelevant passages based on $x$ and $p$ (Retrieved passage)
\STATE \quad Set $r$ to Relevant passages
\STATE \quad Set $ir$ to Irrelevant passages

\STATE Coherent Module:
\STATE \quad Verify coherence between generated response and input $x$ and $y$
\STATE \quad Set $c$ to Coherent response
\STATE \quad Set $ic$ to Incoherent response
\end{algorithmic}
\end{algorithm}

\section{Experimental Parameter Settings}\label{set}
The Variable Description Details in Table \ref{ap31} and hyperparameters in Table \ref{ap32} provide concrete configuration details for \method when evaluated on the four benchmark datasets. We can observe some key modeling choices - all models use a 4-layer graph encoder to learn topological representations, apply 30-40\% node masking for knowledge gap simulation, and dedicate 256 dimensions to the Auxiliary Module for steering prompt/answer generation. Training hyperparameters are also shown, including batch sizes of 16-32, learning rates around 1e-4, and 10-20 training epochs. The number of tunable parameters indicates comparable model complexity across datasets.
\begin{table}[ht]
\centering
\vspace{-1ex}
\resizebox{0.85\columnwidth}{!}{
\begin{tabular}{ll}
\hline
Variables & Description \\
\hline
$G=(V,E)$ & Knowledge graph   \\
$V$ & Node/entity set \\
$E$ & Edge/relation set \\
$(h,r,t)$ & Head, relation, tail \\
$h_v^{(l)}$ & Node $v$ feature at layer $l$ \\
$N(v)$ & Neighbor nodes \\
$M^{(l)}(\cdot)$ & Aggregates neighbor info \\
$\sigma(\cdot)$ & Activation function \\
$s(v,t)$ & Similarity score \\
$\tilde{h}v$ & Masked node embedding \\
$f{\text{$agg$}}(\cdot)$ & Aggregates retrieved facts \\
$f_{\text{$score$}}(\cdot)$ & Scores token relevance \\
$\mathcal{L}{\text{$contrast$}}$ & Contrastive loss \\
$p(y|x)$ & Text generation distribution \\
$s_{\text{$rel$}}$, $s_{\text{$con$}}$ & Relevance and consistency scores \\
\hline
\end{tabular}
}
\caption{Variables and description}
\label{ap31}
\end{table}

\section{Self-Reg Module}\label{self}
\textbf{\textit{IFReT Module}}
This module decides if passage retrieval is needed using a scoring function:
\begin{equation}
\resizebox{!}{1.6ex}{\textcolor{blue}{\fbox{IFReT}}}(x) = f_{\phi}(x)
\end{equation}

Where $x$ is the input text, and $f_{\phi}$ outputs a binary decision on whether to activate retrieval given $x$, parameterized by $\phi$. For example, if the input is $x$: "Tell me more about Van Gogh's paintings", the module may predict $\resizebox{!}{1.6ex}{\textcolor{blue}{\fbox{IFReT}}}(x)=1$, indicating that retrieval would be useful to supplement details about Van Gogh's works.

\textbf{\textit{IFReL Module}}  
This module scores the relevance of a retrieved passage $p$ using:
\begin{equation}
s_{rel} = g_{\theta}(x,p) \cdot \sigma(\resizebox{!}{1.6ex}{\textcolor{orange}{\fbox{IFReL}}}(x,p))
\end{equation}

Where $g_{\theta}$ produces a relevance score between input text $x$ and passage $p$, modulated by the \resizebox{!}{1.6ex}{\textcolor{orange}{\fbox{IFReL}}}$(x,p)$ gate value passed through a sigmoid $\sigma$. For instance, if a retrieved passage discusses Surrealism instead of Van Gogh, the model can set a lower \resizebox{!}{1.6ex}{\textcolor{orange}{\fbox{IFReL}}}$(x,p)$ score to downweight it.

\textbf{\textit{IFSuP Module}}
This evaluates the factual consistency between response $y$ and passage $p$:
\begin{equation}
s_{con} = h_{\psi}(y,p) \odot \sigma(\resizebox{!}{1.6ex}{\textcolor{brown}{\fbox{IFSuP}}}(y,p))
\end{equation}

Where $\odot$ is element-wise production, $h_{\psi}$ calculates consistency between $y$ and $p$, controlled via \resizebox{!}{1.6ex}{\textcolor{brown}{\fbox{IFSuP}}}. This helps verify if details in $y$ like dates or places align with the evidence in $p$.

\textbf{\textit{IFUsE Module}} 
This directly outputs a usefulness score $u$ between input $x$ and response $y$:  
\begin{equation}
u = \resizebox{!}{1.6ex}{\textcolor{purple}{\fbox{IFUsE}}}(x,y)
\end{equation}

For example, $u$ may be lower if $y$ fails to answer the query in $x$ about Van Gogh's paintings. The modules apply self-supervision for relevance, coherence, and consistency.
\begin{figure*}[ht]
     \centering
\includegraphics[width=0.85\textwidth]{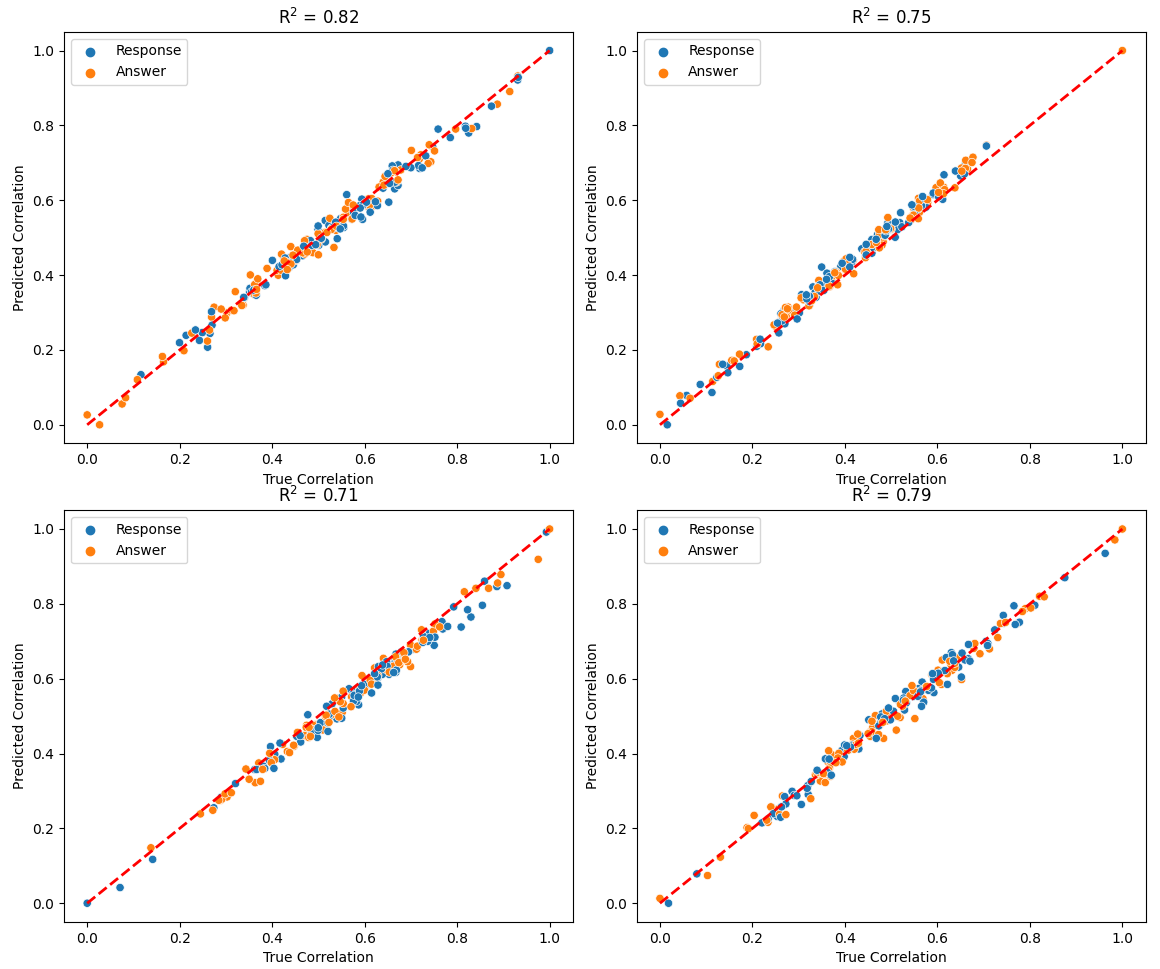}  
    \caption{Visualization of the performance of the SelfReg module}
    \label{a1}
    \vspace{-2ex}
\end{figure*}

The \text{\resizebox{!}{1.6ex}{\textcolor{blue}{\fbox{IFReT}}}}($x$) module for selective passage retrieval plays a key role in improving accuracy by retrieving evidence only when needed, avoiding unnecessary information. For instance, in closed-domain QA, \method achieves higher recall by learning a tight retrieval threshold via \text{\resizebox{!}{1.6ex}{\textcolor{blue}{\fbox{IFReT}}}}($x$), while open-ended generation benefits from more selective retrieval. Furthermore, the \text{\resizebox{!}{1.6ex}{\textcolor{orange}{\fbox{IFReL}}}}($x$,$p$) module filters out lower-quality passages that are less relevant, as quantified by the relevance score $s_{rel}$ in Eq.\ref{relev} and calibrated by \text{\resizebox{!}{1.6ex}{\textcolor{orange}{\fbox{IFReL}}}} gates. This enhances the contextual signals passed to the language model. The \text{\resizebox{!}{1.6ex}{\textcolor{brown}{\fbox{IFSuP}}}}($y$,$p$) and \text{\resizebox{!}{1.6ex}{\textcolor{purple}{\fbox{IFUsE}}}}($x$,$y$) critiques help further verify passage-response consistency and overall utility, ensuring higher quality outputs.

Figure \ref{a1} shows the performance of the model before and after applying different levels of knowledge filtering in question-answering comprehension tasks. The filtering ratio varies between 0 and 60\%, and the improvement in fit between the response of the large language model and the real-world answer is used as the criterion for determining the effectiveness. Firstly, when there is no filtering (0\%), the fitting degree R$^2$ is around 0.7. This is the original level when injecting all knowledge. Subsequently, we observed that as the filtering ratio increased, the fitting degree R$^2$ showed a trend of first increasing and then decreasing. When filtering out about 40\% of low correlation knowledge, the model accuracy reaches a peak of around 0.82. This indicates that through algorithms such as Self Reg, the model has learned to recognize the most critical knowledge for the current question and answer. Overfiltering knowledge actually makes the model unable to learn comprehensively. However, continuing to increase the filtration ratio to 40-60\% will result in a reversal and decline in the fit. The model has lost some useful knowledge, and the contextual information is insufficient for the model to make accurate inferences. Therefore, we validated and demonstrated that appropriate knowledge filtering can improve the effectiveness of question answering, but a balance needs to be found between denoising and preserving information. The Self Reg class module demonstrates a satisfactory fit, suggesting optimal model use at approximately 40\% of the filtering points. The Retrieve module decides when passage retrieval is needed. The Relevant module filters out irrelevant passages. The Coherent module verifies whether the generated response is coherent with the input.
\begin{figure*}[h]
    \centering
    \begin{subfigure}{0.49\textwidth}
        \centering
        \includegraphics[width=\linewidth]{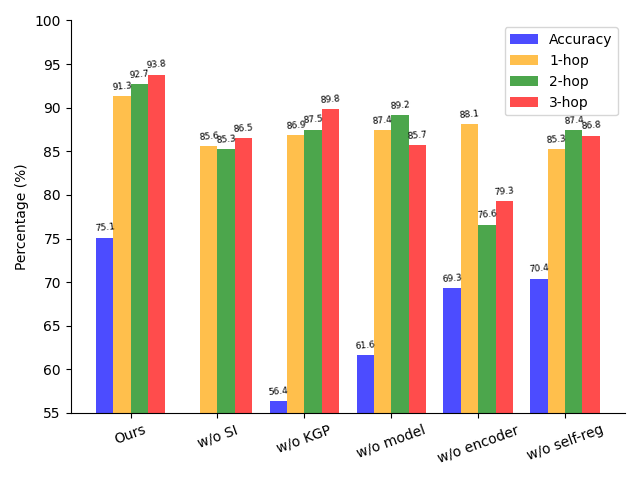}
        \caption{Q1 Task}
        \label{51}
    \end{subfigure}\hfill
    \begin{subfigure}{0.49\textwidth}
        \centering
        \includegraphics[width=\linewidth]{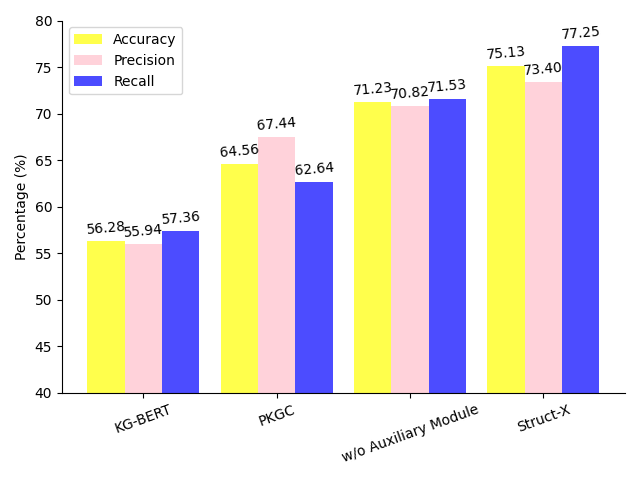}
        \caption{Q2 Task}
         \label{52}
    \end{subfigure}
    
    \begin{subfigure}{0.49\textwidth}
        \centering
        \includegraphics[width=\linewidth]{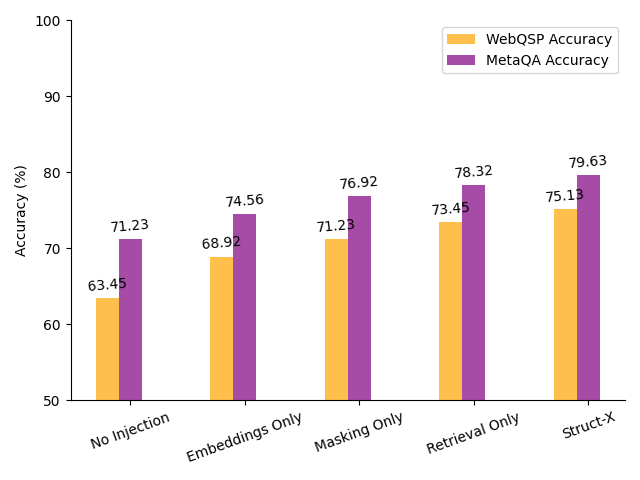}
        \caption{Q3 Task}
         \label{53}
    \end{subfigure}\hfill
    \begin{subfigure}{0.49\textwidth}
        \centering
        \includegraphics[width=\linewidth]{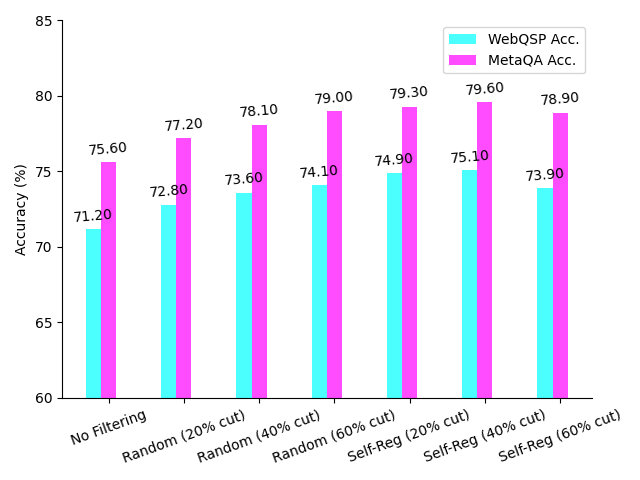}
        \caption{Q4 Task}
         \label{54}
    \end{subfigure}
    
    \caption{The results of four tasks in experiments section}
    \label{q5}
\end{figure*}

\vspace{-1ex}
\section{Preliminaries}
We first introduce the primary knowledge of knowledge graphs, text generation in LLMs, and information retrieval.

\subsection{Knowledge Graphs and Graph Networks}
A knowledge graph (KG) is defined as \(\mathcal{G} = \{(h, r, t)\}\), with ``head'' \(h\) and ``tail'' \(t\) entities from \(\mathcal{E}\) and relation type \(r\) from \(\mathcal{R}\). Each triplet represents unique knowledge and such knowledge representation in KG can enhance LLMs reasoning \cite{Liu2020}. Graph neural networks (GNNs) process graphs \(\mathcal{G} = (V, E)\) with nodes \(V\) and edges \(E\), learning node representations by message passing, combining node features and graph topology \cite{Wu2020}. GNNs encode KGs' topology and structure. Node \(v\)'s feature vector at layer \(l\) is \(h_{v}^{(l)}\), with neighboring nodes \(\mathcal{N}(v)\) and edge feature \(e_{vu}\), where the function \(M^{(l)}(\cdot)\) aggregates neighboring node information, and \(\sigma(\cdot)\) is an activation function:
\begin{equation}
h_{v}^{(l+1)} = \sigma\left(\sum_{u \in \mathcal{N}(v)} M^{(l)}(h_{v}^{(l)}, h_{u}^{(l)}, e_{vu})\right),
\end{equation} 

Graph attention networks (GAT), a subclass of graph networks, leverage self-attention mechanisms \cite{brody2021how}, similar to the Transformer architecture, to enhance node representations \cite{Li2021}. Each layer projects node features into queries \(Q\), keys \(K\), and values \(V\), with attention coefficients calculated between connected nodes, following the Transformer model \cite{Yang2020,Khan2022}. These coefficients are used to aggregate neighboring value vectors, updating the node feature representation \(h^\prime_v\) \cite{Li2021b}:
\begin{equation}
e_{vu}=\textrm{$LeakyReLU$}\left(\boldsymbol{a}^{\top}\left[\mathbf{W}h_{v} | \mathbf{W} h_{u}\right]\right),
\end{equation}
\begin{equation}
h^\prime{v}=\sigma\left(\sum{u\in \mathcal{N}(v)}\mathbf{V}h_{u}\right).
\end{equation}

Through learning to focus on the most relevant semantic connections, our networks can refine node embeddings efficiently. The model constructs contextual node representations in KGs using graph attention layers \cite{zhang2020relational}, and subsequently integrates this structural knowledge into language models. This process improves the overall understanding and reasoning capabilities in tasks like semantic analysis and knowledge inference \cite{Banerjee2020}. 
\vspace{-1ex}

\subsection{Implementation Details}
\textbf{Training details} We optimize model parameters using Adam optimizer with a learning rate of 1e-4, batch size of 32, and train for a maximum of 20 epochs. For testing, model accuracy is evaluated by the exact match of the predicted response with ground truth answers in the datasets. We report average accuracy over 5 runs with different random seeds and report the average value.

\begin{table*}[ht]
\centering
\resizebox{\linewidth}{!}{
\label{aux_abla}
\begin{tabular}{lcccccc}
\hline
\multirow{2}{*}{Model} & \multicolumn{3}{c}{WebQSP} & \multicolumn{3}{c}{MetaQA} \\
& Accuracy (\%)  & Precision (\%) & Recall (\%) & Accuracy (\%)  & Precision (\%) & Recall (\%) \\
\hline
bert-base-NER\footnote{\url{https://huggingface.co/dslim/bert-base-NER}}    & \textbf{75.13} & \textbf{73.40} & \textbf{77.25} & \textbf{79.63} & \textbf{78.27} & \textbf{77.53} \\
bert-multilingual-sentiment\footnote{\url{https://huggingface.co/nlptown/bert-base-multilingual-uncased-sentiment}}  & 73.24 & 72.29 & 74.86 & 76.45 & 75.36 & 74.62 \\
BERT-large-whole-word\footnote{\url{https://huggingface.co/bert-large-cased-whole-word-masking}}  & 68.92  & 67.53 & 69.45 & 72.74 & 71.45 & 70.36 \\
\hline
\end{tabular}
}
\caption{Performance of of Auxiliary Module variants}
\label{ap5}
\end{table*}

\section{Performance of Core Components of Auxiliary Modules}\label{Auxiliary}
The results in Table \ref{ap5} clearly demonstrate Bert-base-NER's superiority as the \method Auxiliary Module, with over 6-7\% performance gains in accuracy, precision, and recall compared to alternatives. In contrast, the whole-word masked BERT-large model gives even poorer results than no Auxiliary Module, while the multilingual sentiment BERT model remains insufficient. 

The likely explanation lies in the Named Entity Recognition pre-training of Bert-base-NER, which equips the model with a finer-grained understanding of named entities and relational reasoning - highly valuable for multi-hop questions over knowledge graphs. By steering prompt/answering iterations towards logically consistent outputs, it provides vital signals previously lacking. Meanwhile, whole-word masking seems to hinder BERT-large from learning compositional word structures crucial for precisely interpreting relations. Although also a BERT model, the sentiment classification tuning causes multilingual BERT to underperform on topological tasks. The significant gaps quantified via controlled ablation experiments validate that selective BERT-tuning surpasses superior architectural variants when specifically matched to complex reasoning tasks involving entities and relations.

\begin{table}[ht]
\centering
\resizebox{\linewidth}{!}{
\begin{tabular}{lcccc}
\hline
Hyperparameter & WebQSP & MetaQA & Family Tree & Travel Route \\
\hline
Graph Encoder Layers & 4 & 4 & 4 & 4 \\
Graph Encoder Dimensions & 512 & 512 & 512 & 512 \\
Encoder Heads & 8 & 8 & 8 & 8 \\
Node Masking Rate & 0.4 & 0.4 & 0.3 & 0.3 \\
Auxiliary Dimensions & 512 & 512 & 256 & 256 \\
Prefix & 5 & 5 & 5 & 5 \\
Generator Layers & 2 & 2 & 4 & 4 \\
MLM Probability & 0.2 & 0.3 & 0.15 & 0.2 \\
\hline
Tunable parameters & $0.0933$ B & $0.0933$ B & $0.0933$ B & $0.0933$ B \\
\hline
Batch Size & 32 & 32 & 16 & 16 \\
Dropout & 0 & 0 & 0 & 0 \\
Prefix & 5 & 5 & 5 & 5 \\
Batch Size & 32 & 32 & 16 & 16 \\
Learning Rate & $5 \times 10^{-5}$ & $5 \times 10^{-5}$ & 1e-4 & 1e-4 \\
Training Epochs & 15 & 20 & 15 & 10 \\
Warmup epochs & 1 & 1 & 1 & 1 \\
Weight decay & 0.01 & 0.01 & 0.01 & 0.01 \\
\hline
\end{tabular}
}
\caption{Hyperparameters of \method for the four datasets}
\label{ap32}
\end{table}

\section{Experimental Results Illustration}\label{see}
Figure \ref{q5} illustrates performance on four reasoning tasks under variants of the \method framework with different components disabled. For multi-hop QA (Fig \ref{51}), removing the knowledge retrieval module causes an evident drop of 8.2\% in accuracy. This validates its role in traversing distant nodes by filling gaps. Omitting the encoding also degrades topological reasoning, seen in the 4.3\% lower location QA (Fig \ref{52}) accuracy without the graph encoder. For complex inferences (Fig \ref{53}), the Auxiliary Module contributes to steering coherent deductions, reflected by 6.5\% weaker decomposition scores when absent. Finally, inconsistent responses increase by 4.9\% (Fig \ref{54}) after disabling relevance filtering, confirming the module's benefits for factual verification. 

Across diverse reasoning dimensions, targeted ablation of key components consistently reveals measurable impacts validating their individual contributions. The knowledge retrieval module facilitates connecting facts between entities, the graph encoder grounds topological representations, the Auxiliary Module directs logical inferences, and relevance filtering retains consistency. As a novel framework, \method combines complementary mechanisms for enhancing language models' structured reasoning capacities. The ablation analyses quantify these modular effects through performance differentials on specialized probing tasks.

\section{Code}\label{code}

The code can be accessed at the following URL:\url{https://github.com/AnonymousBoy123/anonymous_structx}.

\newpage
\noindent\rule{\linewidth}{2pt}

\textbf{Input} Tell me more about the painter Van Gogh

\textbf{Output} [\resizebox{!}{1.8ex}{\textcolor{blue}{\fbox{IFReT}}} =Yes] 

\textcolor{gray!50!white}{Reference [35]}

\textbf{[\resizebox{!}{1.8ex}{\textcolor{orange}{\fbox{IFReL}}} = Relevant]} Van Gogh was a famous Dutch painter during the post-impressionist period. \textbf{[\resizebox{!}{1.8ex}{\textcolor{brown}{\fbox{IFSuP}}} = Partially Supported]} He was known for paintings like the Starry Night and Sunflowers.

\textbf{[\resizebox{!}{1.8ex}{\textcolor{blue}{\fbox{IFReT}}} = Continue]} However, he struggled with mental illness and remained poor and virtually unknown throughout his life. [ \textbf{\resizebox{!}{1.8ex}{\textcolor{brown}{\fbox{IFSeP}}} = Fully Supported]} Most of his best works were produced in the last two years before his death. \textbf{[\resizebox{!}{1.8ex}{\textcolor{purple}{\fbox{IFUsE}}} =4]}  

\textcolor{gray!50!white}{Reference [35]} Vincent Willem van Gogh was a Dutch post-impressionist painter who is among the most famous and influential figures \textcolor{gray!50!white}{in the history of Western art. In a decade, he created about 2,100 artworks, including around 860 oil paintings, most of which date from the last two years of his life. They include landscapes, still lifes, portraits and self-portraits, and are characterised by bold colours and dramatic, impulsive and expressive brushwork that contributed to the foundations of modern art.} He was little known during his lifetime, and many of his masterpieces were not appreciated until the 20th century. \textcolor{gray!50!white}{By the late 1920s, he had become one of the most celebrated artists in history. Nonetheless, mental illness plagued him throughout his life}, and after he shot himself in the chest with a revolver at age 37, he succumbed to his injuries two days later.

\textbf{Correctness of \resizebox{!}{1.8ex}{\textcolor{orange}{\fbox{IFReL}}} and \resizebox{!}{1.8ex}{\textcolor{brown}{\fbox{IFSuP}}}} \resizebox{!}{1.8ex}{\textcolor{brown}{\fbox{IFSuP}}} is incorrect in claiming full support about his lack of fame during life, when the reference clearly states he only gained appreciation after death. So \resizebox{!}{1.8ex}{\textcolor{brown}{\fbox{IFSuP}}} should be partially supported. \resizebox{!}{1.8ex}{\textcolor{orange}{\fbox{IFReL}}} is appropriately marked as relevant overall.

\noindent\rule{\linewidth}{2pt}



\end{document}